\documentclass[sigconf]{acmart}
\AtBeginDocument{%
  }

\copyrightyear{2025}
\acmYear{2025}
\setcopyright{acmlicensed}
\acmConference[MM '25]{Proceedings of the 33rd ACM International Conference on Multimedia}{October 27--31, 2025}{Dublin, Ireland}
\acmISBN{978-1-4503-XXXX-X/xxxx/xx}
\acmDOI{XXXXXXX.XXXXXXX}

\usepackage{subcaption}
\usepackage{graphicx}
\usepackage{multirow}
\usepackage{makecell}
\usepackage{array}

\begin{document}

\title{Disentangling Homophily and Heterophily in Multimodal Graph Clustering}


\author{Zhaochen Guo}
\authornotemark[1]
\affiliation{%
  \institution{University of Electronic Science and
Technology of China}
  \city{Chengdu}
  \state{Sichuan}
  \country{China}
}
\email{2023310101003@std.uestc.edu.cn}

\author{Zhixiang Shen}
\authornote{Both authors contributed equally to this research.}
\affiliation{%
  \institution{University of Electronic Science and
Technology of China}
  \city{Chengdu}
  \state{Sichuan}
  \country{China}
}
\email{zhixiang.zxs@gmail.com}

\author{Xuanting Xie}
\affiliation{%
  \institution{University of Electronic Science and
Technology of China}
  \city{Chengdu}
  \state{Sichuan}
  \country{China}
}
\email{x624361380@outlook.com}

\author{Liangjian Wen}
\affiliation{%
  \institution{Southwestern University of Finance and Economics}
  \city{Chengdu}
  \state{Sichuan}
  \country{China}
}
\email{wlj6816@gmail.com}

\author{Zhao Kang}
\authornote{Corresponding author}
\affiliation{%
  \institution{University of Electronic Science and
Technology of China}
  \city{Chengdu}
  \state{Sichuan}
  \country{China}
}
\email{zkang@uestc.edu.cn}



\renewcommand{\shortauthors}{Zhaochen Guo, Zhixiang Shen et al.}
\sloppy
\begin{abstract}
Multimodal graphs, which integrate unstructured heterogeneous data with structured interconnections, offer substantial real-world utility but remain insufficiently explored in unsupervised learning. In this work, we initiate the study of multimodal graph clustering, aiming to bridge this critical gap. Through empirical analysis, we observe that real-world multimodal graphs often exhibit hybrid neighborhood patterns, combining both homophilic and heterophilic relationships. To address this challenge, we propose a novel framework—Disentangled Multimodal Graph Clustering (DMGC)—which decomposes the original hybrid graph into two complementary views: (1) a homophily-enhanced graph that captures cross-modal class consistency, and (2) heterophily-aware graphs that preserve modality-specific inter-class distinctions. We introduce a Multimodal Dual-frequency Fusion mechanism that jointly filters these disentangled graphs through a dual-pass strategy, enabling effective multimodal integration while mitigating category confusion. Our self-supervised alignment objectives further guide the learning process without requiring labels. Extensive experiments on both multimodal and multi-relational graph datasets demonstrate that DMGC achieves state-of-the-art performance, highlighting its effectiveness and generalizability across diverse settings. Our code is available at https://github.com/Uncnbb/DMGC.

\end{abstract}

\ccsdesc[500]{Computing methodologies~Cluster analysis}
\ccsdesc[300]{Computing methodologies~Neural networks}
\ccsdesc[300]{Computing methodologies~Unsupervised learning}

\keywords{Unsupervised multimodal learning, Multimodal clustering, Multiplex graph, Multimodal fusion and alignment}


\maketitle

\section{Introduction}

\begin{figure*}[h]
\centering
\subfloat[Amazon (Multimodal Multi-relational Graph) \label{Amazon-1D}]{
    \includegraphics[width=0.42\columnwidth]{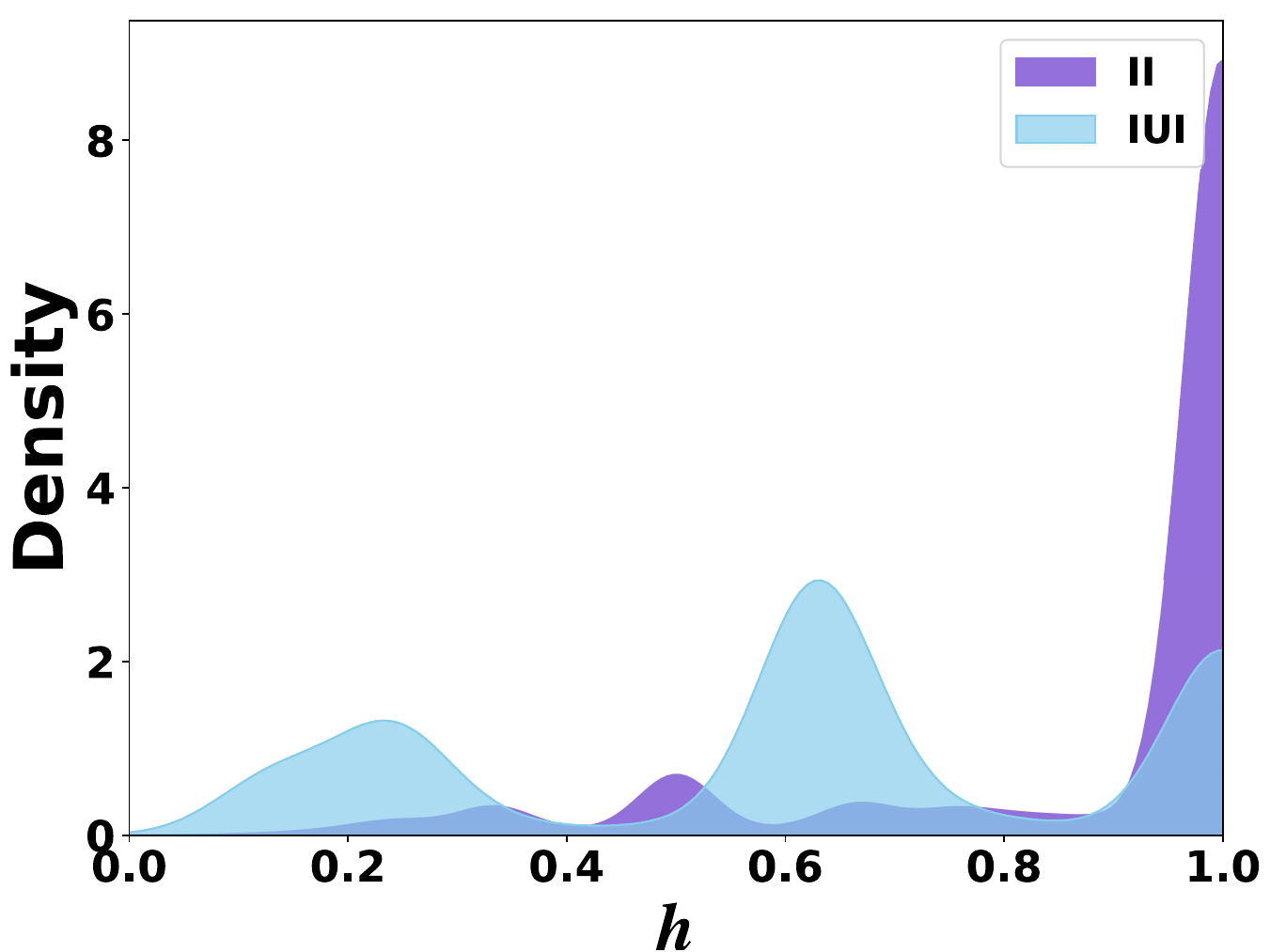}
    \label{fig:Amazon}
}
\hspace{2em}
\subfloat[IMDB (Multimodal Multi-relational Graph)\label{IMDB-1D}]{
    \includegraphics[width=0.42\columnwidth]{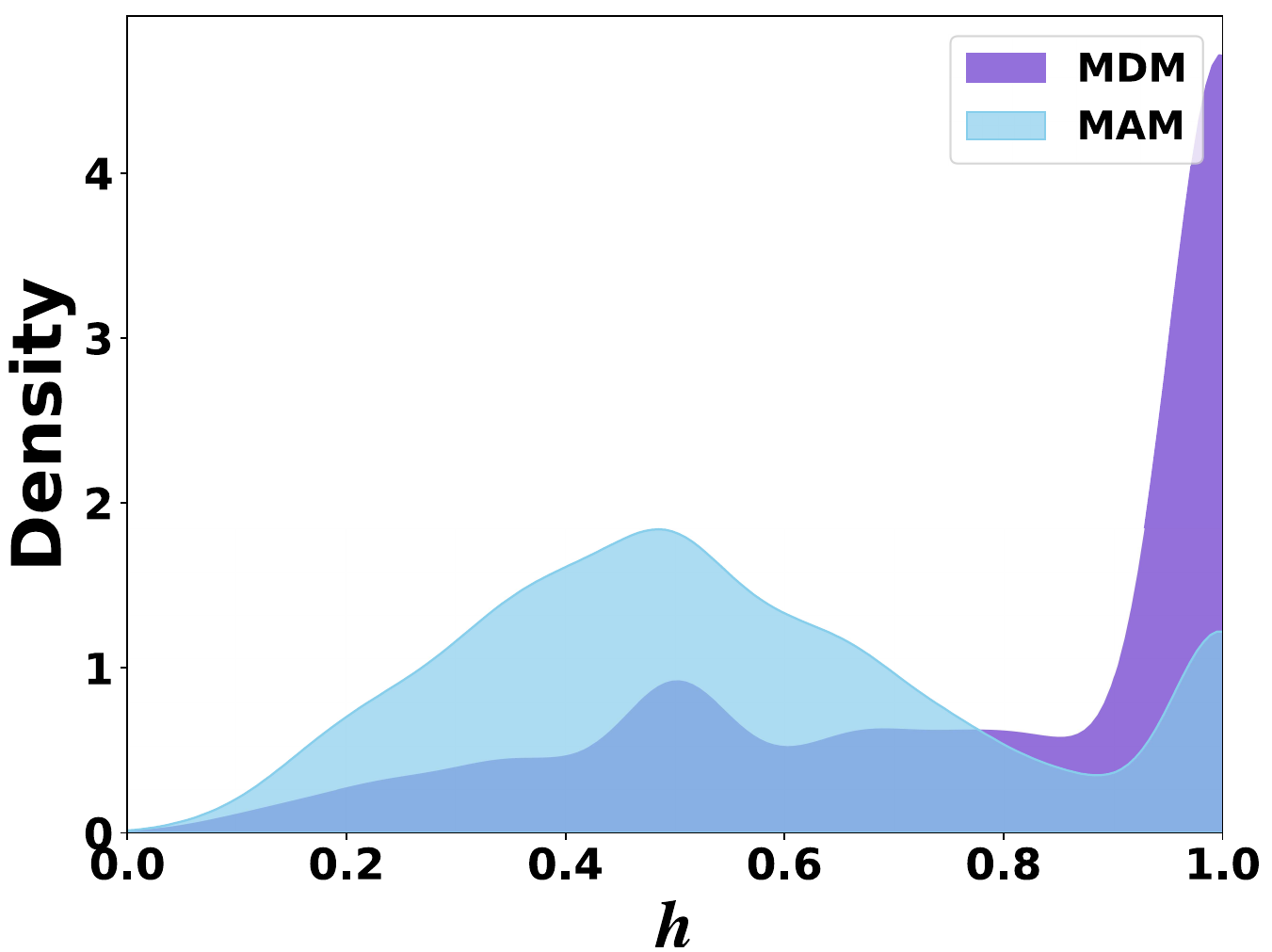}
    \label{fig:IMDB}
}
\hspace{2em}
\subfloat[Ele-Fashion (Multimodal Graph)\label{Ele fashion-1D}]{
    \includegraphics[width=0.42\columnwidth]{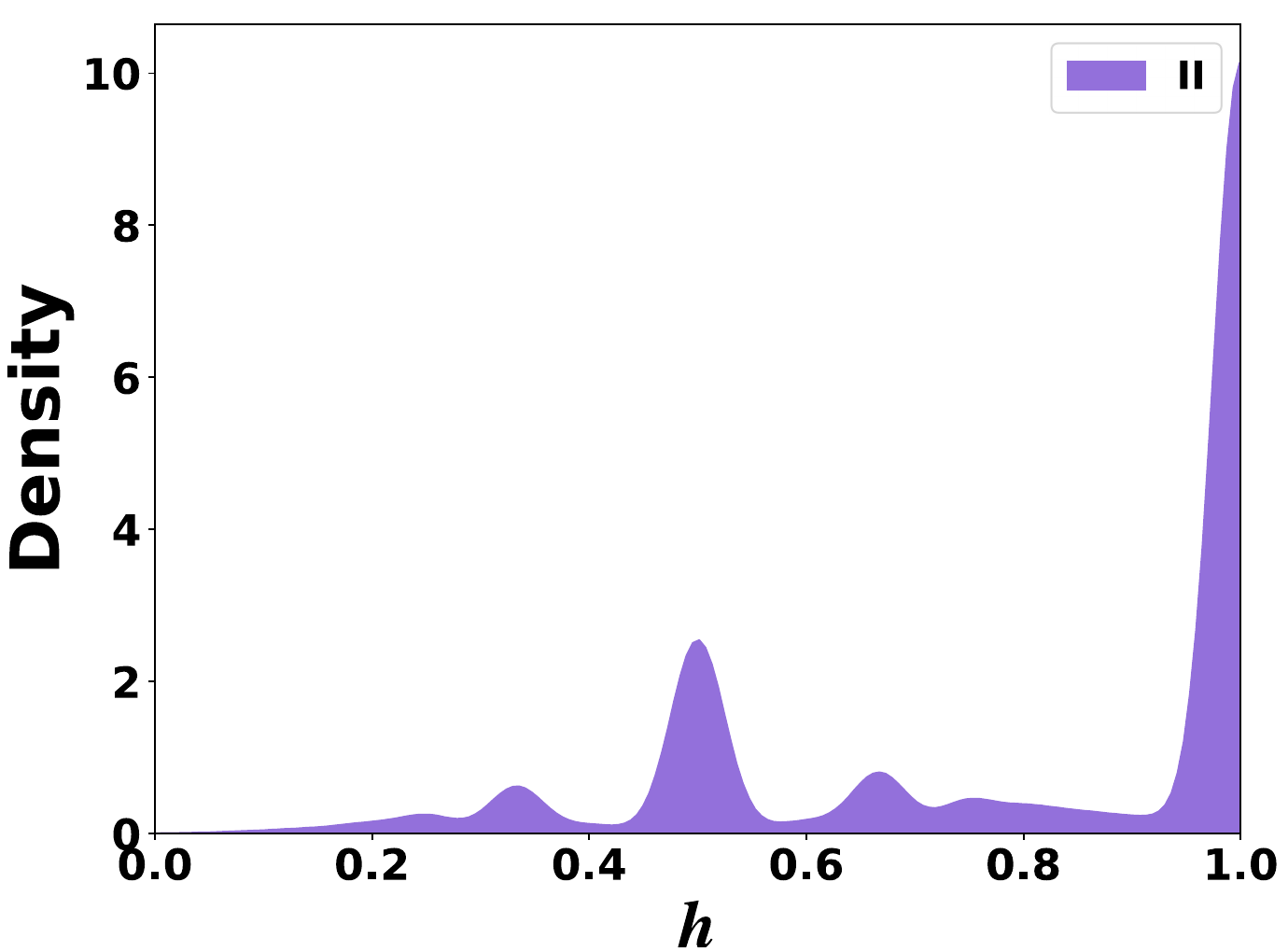}
    \label{fig:Ele fashion}
}
\hspace{2em}
\subfloat[Yelp (Multi-relational Graph) \label{Yelp-1D}]{
    \includegraphics[width=0.42\columnwidth]{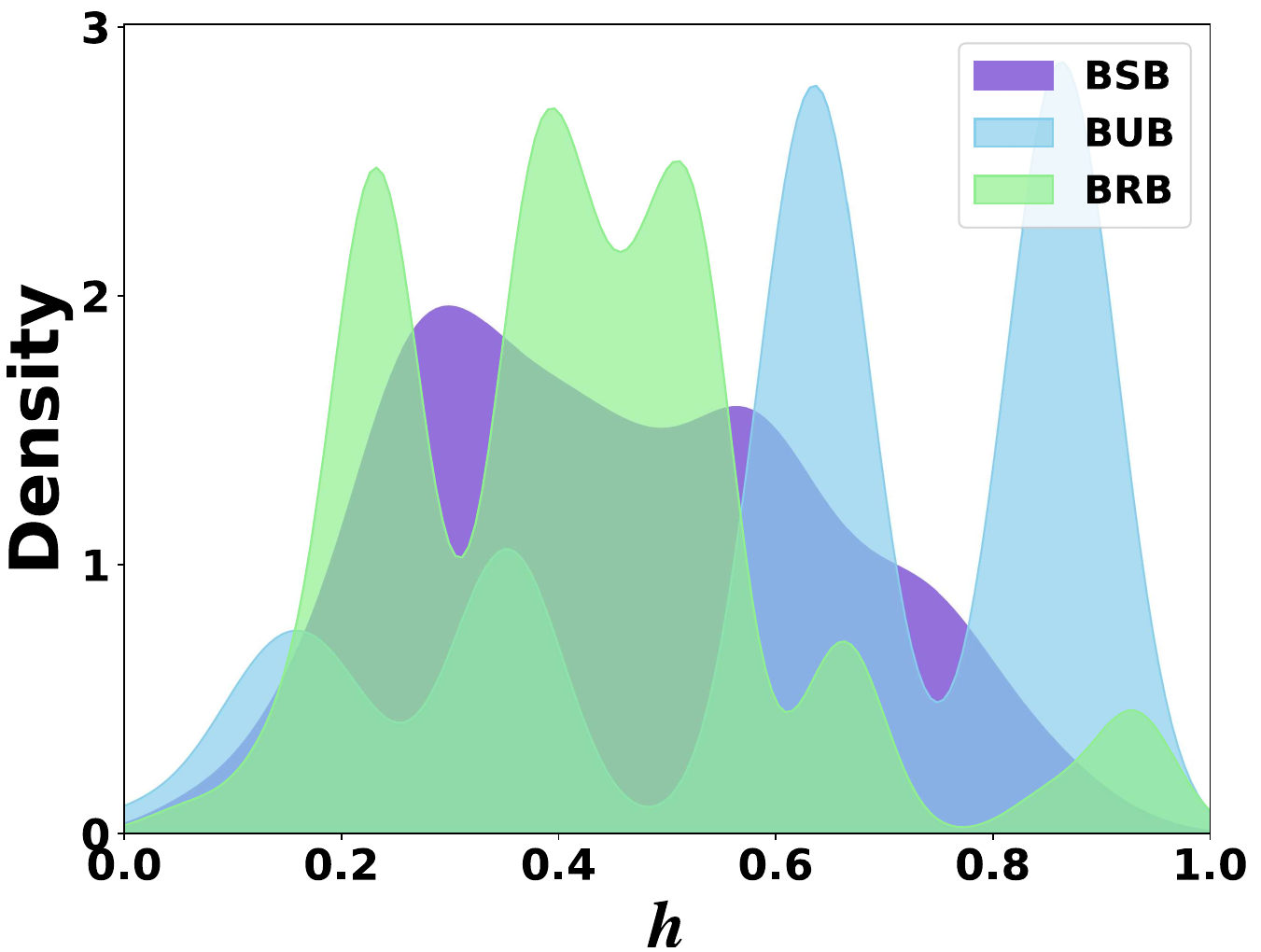}
    \label{fig:yelp}
}

\caption{Node-level homophily ratio distributions across different datasets reveal a prevalent {\textit{hybrid neighborhood pattern}}. Many nodes exhibit moderate homophily levels, where their neighborhoods blend both homophilic and heterophilic neighbors.}
\Description{Visualization of node-level homophily ratio distributions in multimodal and multi-relational graphs, illustrating the presence of many low-homophily nodes.}
\label{Empirical_Research}
\end{figure*}

Multimodal graphs—integrating unstructured heterogeneous modalities such as text, images, and audio with structured topological connections—have found widespread applications across diverse domains, including social network analysis \cite{wang2023tiva}, recommendation systems \cite{tao2020mgat}, and bioinformatics \cite{li2024multimodal} \cite{ektefaie2023multimodal,peng2024learning}. These graphs offer a rich representation by capturing both the content and structural relationships among entities, making them particularly powerful for real-world modeling.

Multimodal learning, as a pivotal advancement in graph representation learning, has driven progress in core tasks such as node classification \cite{he2025unigraph2}, link prediction \cite{li2023imf}, and knowledge graph completion \cite{chen2022hybrid}. However, these tasks typically rely on large quantities of high-quality annotated data \cite{zhu2024multimodal}, which are often costly or impractical to obtain in many real-world scenarios. To address this limitation, we pioneer the systematic study of multimodal graph clustering—a critical yet largely underexplored unsupervised learning task that facilitates deeper structural and semantic understanding of complex multimodal graphs.

Our work departs significantly from conventional graph-based multimodal clustering approaches \cite{xia2023graph}, which generally operate on unstructured data (e.g., feature matrices) without leveraging the full potential of graph topologies. In contrast, we focus on the more challenging and practical problem of clustering over multimodal graph data, where both the structural dependencies and modality-specific information must be jointly exploited. This setting introduces unique challenges but also opens the door to more expressive and generalizable models for unsupervised learning.

Despite recent advancements, a fundamental challenge in modeling multimodal graphs lies in their inherently complex connectivity structures. Real-world graphs often contain diverse edge types \cite{jia2023multimodal,chen2021graph,shen2024balanced}, each capturing unique relational semantics across and within modalities. However, existing methods largely operate under the single-relational assumption \cite{xie2024one,xia2023graph,yan2023gcfagg}, which significantly limits their ability to model the rich, modality-aware interactions present in realistic scenarios.

This limitation becomes especially critical in unsupervised settings, where the absence of labeled data necessitates the effective extraction of self-supervised signals from complex multi-relational patterns \cite{shen2024beyond,jing2021hdmi}. Without such mechanisms, clustering performance suffers due to inadequate structural and semantic alignment across modalities.

To address these challenges, we propose a unified framework that jointly performs multimodal feature fusion and multi-relational structure learning, enabling a more comprehensive and discriminative representation of multimodal graphs. Our approach lays the groundwork for a more holistic understanding of graph data by bridging the gap between feature integration and relational diversity.

Beyond the complexity of multi-relational structures, the presence of heterophilic connections in multimodal graphs introduces an additional challenge that further undermines structural reliability. As shown in Figure \ref{Empirical_Research}, the Gaussian Kernel Density Estimation \cite{terrell1992variable} of Node-level Homophily Ratio \cite{luan2022revisiting} indicates a broad spectrum of values, which aligns with the characteristic hybrid neighborhood pattern found in real-world multimodal graphs, where nodes typically maintain connections with both homophilic (class-consistent) and heterophilic (class-divergent) neighbors \cite{luan2022revisiting,luan2024heterophilic,10905047}. This results in a moderate homophily level, empirically identified as one of the most difficult scenarios for conventional GNN architectures, which struggle to reconcile conflicting message-passing signals in such settings \cite{chien2020adaptive, chen2023polygcl, pan2023beyond}. Consequently, the effectiveness of existing graph-fixed methods in multimodal graph learning is fundamentally limited.

Motivated by these observations, our research addresses a central question:
\textit{How can we effectively extract and leverage meaningful patterns from raw multimodal graphs characterized by complex multi-relational structures and hybrid neighborhoods, thereby enabling effective multimodal graph clustering?}

To this end, we propose the Disentangled Multimodal Graph Clustering (DMGC) framework. Our approach consists of four core components:

First, we decouple the hybrid multi-relational graph into two complementary views: A cross-modality homophilic graph that reinforces inter-modality consensus by identifying reliable, class-consistent neighbors. Modality-specific heterophilic graphs that preserve diverse, modality-dependent semantic information. Second, we introduce a dual-pass graph filtering mechanism that captures both intra-class commonalities and inter-class distinctions, facilitating more expressive semantic fusion across modalities. Finally, we design novel alignment objectives that mine informative self-supervised signals from both frequency domains and modalities, guiding the learning process without reliance on labels. In summary, our main contributions are three-fold:

\begin{enumerate}
    \item \textbf{Problem}: We present the first systematic investigation of multimodal graph clustering under unsupervised settings, addressing the dual challenges of multi-relational complexity and hybrid neighborhood structures. To the best of our knowledge, this work pioneers principled learning on raw multimodal graph data without supervision.
    
    \item \textbf{Algorithm}: We propose DMGC, a novel framework that integrates disentangled graph construction to separate homophilic and heterophilic patterns, and introduces a multimodal dual-frequency fusion and alignment mechanism to extract rich semantic representations and facilitate effective self-supervised clustering.
    
    \item \textbf{Evaluation}: We perform comprehensive experiments on multiple real-world multimodal and multi-relational graph datasets, benchmarking against various state-of-the-art approaches. The results consistently demonstrate the effectiveness and robustness of our proposed method.  
\end{enumerate}

\section{Related Work}

\subsection{Multimodal Clustering}

Multimodal learning aims to learn unified high-quality representations from heterogeneous modalities \cite{radford2021learning,girdhar2023imagebind,wang2025informationcriterioncontrolleddisentanglement}, including text, images, and audio. Within this paradigm, Multimodal Clustering (MMC) has attracted considerable attention as it seeks to uncover latent patterns in multimodal data through unsupervised learning. GWMAC \cite{10.1145/3511808.3557339} enhances clustering performance by leveraging complementary information across modalities through weighted aggregation. GECMC \cite{xia2023graph}, GAMMC \cite{DBLP:journals/tkde/HeWGWHY24} and CoE \cite{wangcooperation} adopt a modality-specific graph construction strategy to heuristically capture heterogeneous information from different modalities. In contrast, GCFAgg \cite{yan2023gcfagg} learns a global and cross-modality consensus graph to achieve modality alignment.
While existing MMC methods primarily focus on constructing modality-specific or consensus graphs to explore sample correlations, they intrinsically operate on unstructured data and overlook the diverse relations inherent in multimodal data. 

\begin{figure*}[t!]
    \centering
    \includegraphics[width=.98\textwidth]{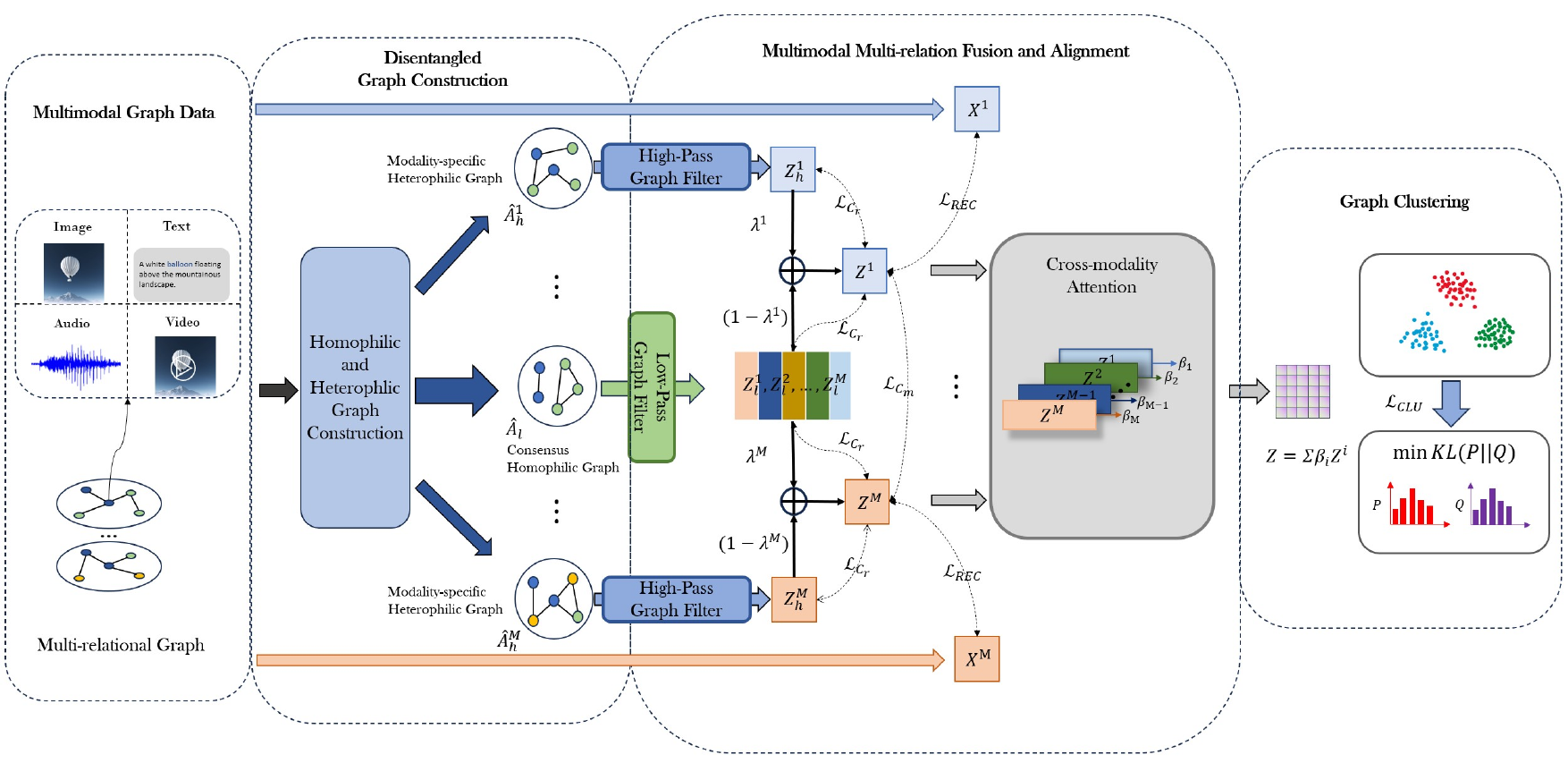}
    \caption{The overall framework of our proposed DMGC. First, it disentangles a cross-modality consensus homophilic graph and \(M\) instances of modality-specific heterophilic graphs through the given multimodal graph data. Subsequently, it performs frequency-domain alignment and modality alignment on the graph-filtered representations, thereby obtaining high-quality fused representations suitable for clustering tasks.}
    \label{fig:pipline}
\end{figure*}

\subsection{Multimodal Graph Learning}

Multimodal Graph Learning (MMGL) seeks to jointly leverage both unstructured heterogeneous modalities and structured topological connections \cite{zhu2024multimodal,yan2024graph}, demonstrating successful applications across diverse domains such as multimodal generation \cite{yoon2023multimodal}, recommendation systems \cite{tao2020mgat}, and molecular optimization \cite{jin2018learning}. Representative approaches include MGAT \cite{tao2020mgat}, which employs attention mechanisms to model multimodal interaction graphs for enhanced user behavior understanding in recommendation systems. MHGAT \cite{jia2023multimodal} further considers diverse types of relations across all modalities. Meanwhile, UniGraph2 \cite{he2025unigraph} establishes a unified multi-domain foundation model for multimodal graphs and achieves versatile task adaptation through knowledge transfer. Despite these advancements, current MMGL research predominantly focuses on supervised learning scenarios, leaving multimodal graph clustering as a critical yet systematically underexplored frontier.

Early methods like MvAGC \cite{lin2023multi} and MCGC \cite{pan2021multi} integrate graph filtering with unsupervised techniques (e.g., spectral clustering) to uncover latent patterns in multi-relational graphs. The advent of deep representation learning shifts focus toward graph neural networks (GNNs) \cite{wu2020comprehensive} and self-supervised learning \cite{liu2021self}. O2MAC \cite{fan2020one2multi} introduces GNNs for multi-relational graph reconstruction to capture shared information. Subsequent approaches like DMGI \cite{park2020unsupervised} and HDMI \cite{jing2021hdmi} optimize mutual information across relational contexts to fuse multi-view representations. Recent methods such as MGDCR \cite{mo2023multiplex}, BTGF \cite{qian2024upper}, and BMGC \cite{shen2024balanced} employ contrastive learning to align multi-relational representations. DMG \cite{mo2023disentangled} enhances graph consistency and complementarity through disentangled learning, while InfoMGF \cite{shen2024beyond} takes a structure learning perspective, refining graph structures to eliminate task-irrelevant noise and maximize task-relevant information across shared and unique relations. Although there has been significant progress, most current methods for multi-relational graphs cannot effectively model multimodal information.

\section{Methodology}
As illustrated in Figure \ref{fig:pipline}, our proposed DMGC consists of three modules: the \textit{Disentangled Graph Construction module}, \textit{Multimodal Dual-frequency Fusion module}, and \textit{Alignment Guided Graph Clustering module}.

\textbf{Notation.} The multimodal multi-relational graph is defined as \( \mathcal{G} = (\mathcal{V}, \mathcal{A}, \mathcal{X}) \), where \( \mathcal{V} \) is the node set with \( |\mathcal{V}| = N \), \( \mathcal{A} = \{ A_1, A_2, \dots, A_R \} \) is a set of symmetric adjacency matrices representing \( R \) different relations, and \( \mathcal{X} = \{ X^{1}, X^{2}, \dots, X^{M} \} \) denotes the multimodal node feature set with \( M \) modalities. Specifically, \( X^i \in \mathbb{R}^{N \times d_f} \) is the feature matrix of the \( i \)-th modality, where all features are extracted and projected to a shared dimensionality \( d_f \) via pretrained encoders (e.g., CLIP~\cite{radford2021learning}). \( C \) denotes the number of classes.

\subsection{Disentangled Graph Construction}

\subsubsection{Graph Aggregation}\label{graph_agg}
By filtering the raw features of each modality over adjacency matrices, we derive aggregated features for each modality. Similar to the existing graph convolution \cite{gasteiger2018predict}, we introduce the aggregated features as follows:

\begin{equation}
X^{i(0)}_{r} = X^i, \quad
X^{i(t+1)}_{r} = (1 - \alpha) \tilde{A}_r X^{i(t)}_{r} + \alpha X^i,
\end{equation}
where $\tilde{A}_r$ denotes the normalized $A_r$ and $X^{i(t)}_{r}$ represents the aggregated feature matrix for modality $i$ after $t$-order filtering under the $r$-th relation. The hyperparameter $\alpha$ controls the contribution of the initial input features $X^i$ at each propagation step, effectively acting as a residual connection. Here, $t \in [0, T - 1]$, and the final aggregated features are given by $X_{r,agg}^i = X^{i(T)}_{r}$. By concatenating them together, we have \(\tilde{X^i} = \left[ X^i_{1,agg}, X^i_{2,agg}, \dots, X^i_{R,agg} \right] \in \mathbb{R}^{N \times Rd_f}\).

\subsubsection{Homophilic and Heterophilic Graph Construction}
Real-world graphs often exhibit a prevalent hybrid neighborhood pattern characterized by different homophily. To comprehensively model these dual characteristics, we propose a disentangled graph construction approach that explicitly differentiates homophilic and heterophilic relations.

Given the fact that nodes with both similar neighbors and features often share the same category, we construct a consensus cross-modality similarity matrix \(S\) with strongly related neighbors as follows:

\begin{equation}
\tilde{X} = \sum_{i=1}^{M} w_i \tilde{X}^i, \quad
{S} = \tilde{X} \tilde{X}^\top,
\end{equation}
where \(w_i\) is the weight hyperparameter to each modality \(i\), coarsely adjusting the influence of different modalities on the fused cross-modality feature.

Besides, to extract modality-specific diverse relationships, we construct a heterophilic graph for every modality, which simultaneously preserves the uniqueness. The \(i\)-th intra-modality similarity matrix is \({S}^i = \tilde{X}^i \tilde{X}^{i\top}\). We use the complementary graph of \({S}^i\) to construct a heterophily graph. Then, we apply \(\text{Top}(\cdot, k)\) to select the \( k \) nearest neighbors to ensure a balance between common and modality-specific information:
\begin{equation}
\begin{aligned}
A_l = \text{Top}(\tilde{S}, k_l), \quad
A^i_h = \text{Top}(\mathbf{1.} - \tilde{S}^i, k_h) \\
\hat{A}_l = \tilde{D}_l^{-1/2} \tilde{A}_l \tilde{D}_l^{-1/2}, \quad
\hat{L}^i_h = I - \tilde{D}_h^{-1/2} \tilde{A}^i_h \tilde{D}_h^{-1/2}
\end{aligned}
\end{equation}
where $\tilde{S}$ and $\tilde{S}^i$ are standardized similarity matrices by Min-Max normalization. Besides, \(\tilde{A}\) denotes the matrix \(A\) with the self-loop. \(k_l\) and \(k_h\) are hyperparameters that determine the number of the nearest neighbors sampled from the constructed homophilic and heterophilic graphs, respectively.

\subsubsection{Scalable Approaches for Large-scale Graphs}
Large-scale graph construction faces inherent scalability limitations due to the quadratic complexity of explicit similarity computation. We propose a linear method to build a homophilic graph using locality-sensitive hashing and heterophilic graphs employing anchor-based construction. 

\textbf{Locality-sensitive hashing for homophilic graph.} 
We apply K-means clustering on the  fused feature matrix \( \tilde{X} \) to generate pseudo-labels \(\hat{y}\). Based on this, we construct the homophilic graph by selecting reliable neighbors within each cluster through scalable $k$NN with a locality-sensitive hashing technique, ensuring linear complexity \cite{fatemi2021slapsselfsupervisionimprovesstructure}.

\textbf{Anchor-based construction for heterophilic graph.} For heterophilic graph construction, we introduce modality-specific pseudo-labels that are generated separately for each modality. To identify heterophilic relationships, anchor nodes are selected within each cluster by determining the \(k\) nearest nodes to the cluster center based on Euclidean distance.

The modality-specific similarity matrix \( S_b^i \) is constructed by calculating the pairwise similarity between all nodes and the anchors: 

\begin{equation}
S_b^i = \tilde{X}^i \tilde{X}_{\mathtt{A}}^{i\top} \in\mathbb{R}^{N \times B}
\end{equation}
where \( \tilde{X}^i_{\mathtt{A}} \) is the selected anchor nodes in the \(i\)-th modality. Homophilic and heterophilic graphs can be obtained following the same procedure described in the previous section.

\textbf{Complexity Analysis:}
Due to the sparsity of real-world graphs, we use sparse matrix techniques for graph aggregation in Section \ref{graph_agg}. For simplicity, $B$ represents both the batch size in scalable $k$NN and the number of anchors for scalable graph construction. The time complexity of disentangled graph construction is $\mathcal{O}\left(MRD(ET+NB)\right)$, where the first term corresponds to graph aggregation and the second to scalable graph construction.

\subsection{Multimodal Dual-frequency Fusion}


\subsubsection{Dual-frequency Graph Filtering} 

The process begins by mapping intra-modality aggregated features into a shared latent space via a parameterized multi-relational encoder:

\begin{equation}
{Z}^i_r = f_{\Theta_r}(\tilde{X}_{r,agg}^{i}), \quad r \in \{1, \dots, R\}
\end{equation}
where \( f_{\Theta_r} \) denotes the relation-specific projection function, with parameters \(\Theta_r\) learned independently for each relation \(r\).

The low-pass filter accentuates homophilic structures, reinforcing shared semantics, while the high-pass filter amplifies heterophilic signals, capturing distinctive variations. We separately apply low-pass filtering through the homophilic graph \( \hat{A}_l \) and high-pass filtering via the heterophilic laplacian \( \hat{L}_h^i \):

\begin{equation}
{Z}_{r,l}^{i(0)} = {Z}_{r,h}^{i(0)} = {Z}^i_r
\end{equation}

\begin{equation}
{Z}_{r,l}^{i(\mathbb{k}+1)} = \sigma\left( \hat{A}_l Z_{r,l}^{i(\mathbb{k})} W \right), \quad
{Z}_{r,h}^{i(\mathbb{k}+1)} = \sigma\left( \hat{L}_h^i Z_{r,h}^{i(\mathbb{k})} W \right)
\end{equation}
where \( {Z}_{r,l}^{i(\mathbb{k})} \) and \( {Z}_{r,h}^{i(\mathbb{k})} \) denote the low-pass and high-pass filtered representations at the \( \mathbb{k} \)-th step \(\mathbb{k} \in [0, \mathbb{K} - 1]\), respectively.  The final filtered representations are given by \( {Z}^i_{r,l} = {Z}_{r,l}^{i(\mathbb{K})} \) and \( {Z}^i_{r,h} = {Z}_{r,h}^{i(\mathbb{K})} \). \( \sigma \) is the activation function, \( W \) represents the network parameters and \( {Z}^i_r \) is the initial representation.

For each relation, we obtain the low-pass and high-pass filtered representations, and compute the averaged representations as follows:  \({Z}^i_{l} = \frac{1}{R} \sum_{r=1}^{R} {Z}^i_{r,l}, \ {Z}^i_{h} = \frac{1}{R} \sum_{r=1}^{R} {Z}^i_{r,h}\). The adaptive fusion of dual-frequency representations is formulated as:

\begin{equation}
{Z}^i = \gamma^i {Z}^i_{l} + (1 - \gamma^i) {Z}^i_{h}
\end{equation}
where \( \gamma^i \) is the trainable modality-specific coefficient that dynamically adjusts the balance between homophilic and heterophilic representations.   

\subsubsection{Multimodal Fusion}

We propose an attention-based mechanism to fuse modality-specific representations into a unified latent space. This approach dynamically highlights salient cross-modal information, maintaining structural consistency and enhancing semantic complementarity. Specifically, we apply a shared linear transformation to modality features, followed by a \( \tanh \) activation and learned parameters:

\begin{equation} 
\begin{aligned} 
\omega_i &= \frac{1}{N} \sum_{k=1}^{N} \hat{q}_i^{\top} \cdot \tanh(W \cdot {Z}^{i}_k + b), \quad i \in \{1, 2, \dots, M\}
\end{aligned} 
\end{equation} 

\begin{equation} 
\beta_i = \frac{\exp(\omega_i)}{\sum_{j=1}^{M} \exp(\omega_j)}
\end{equation}  
where \( \omega_i \) represents the relevance score for each modality \( i \), influenced by the learnable attention vectors \( \hat{q}_i \). It adaptively regulates the contributions of each modality, ensuring a modality-aware and semantically enriched fused representation \( Z \) for effective multimodal learning:

\begin{equation} 
{Z} = \sum_{i=1}^{M} \beta_i \cdot {Z}^{i}
\end{equation}

\subsection{Alignment Guided Graph Clustering}


\subsubsection{Intra-modality Reconstruction}

The low-pass and high-pass filtered representations \( {Z}^i_l \) and \( {Z}^i_h \) are concatenated and passed through a decoder, and the reconstructed features are compared with the original ones using cosine similarity. Define \( \hat{X}^i \) = \( g([{Z}^i_{l} \parallel {Z}^i_{h}]) \), where \( g \) is an MLP decoder. The reconstruction loss is defined as:

\begin{equation}
\mathcal{L}_{REC} = \frac{1}{M \cdot N} \sum_{i=1}^{M} \sum_{k=1}^{N} \left( 1 - \frac{{X}_{k}^{i\top}  \hat{X}_{k}^i}{\|{X}_{k}^i\| \|\hat{X}_{k}^i\|} \right),
\end{equation}

\subsubsection{Dual-frequency Alignment}

The multichannel contrastive loss aligns the low-pass and high-pass representations with the fused representation \( {Z}^i \) while preserving their distinct features as follows:

\begin{equation}
\mathcal{L}_{Cr} = \sum_{i=1}^{M}(\mathcal{L}_{\text{contrast}}({Z}^i_l, {Z}^i) + \mathcal{L}_{\text{contrast}}({Z}^i_h, {Z}^i)).
\label{eq:16}
\end{equation}
Same as \cite{wu2018unsupervised}, $\mathcal{L}_{\text{contrast}}$ is defined as the InfoNCE loss applied to node representations.

\subsubsection{Cross-modality Alignment}

While the multichannel contrastive loss aligns modality-specific features with the fused representations, it does not explicitly enforce consistency across different modalities. To address this, we introduce a cross-modality alignment loss that brings modality-specific but semantically similar representations closer in the shared embedding space, promoting a unified representation for cross-modality understanding and retrieval:

\begin{equation}
\mathcal{L}_{Cm} = \sum_{i=1}^{M} \sum_{j=i+1}^{M} \mathcal{L}_{\text{contrast}}(Z^i, Z^j).
\end{equation}

\begin{equation}
\begin{split}
\mathcal{L}_{\text{contrast}}(Z^i, Z^j) = \frac{1}{2N} \sum_{k=1}^N \big(\ell(Z_k^i, Z_k^j) + \ell(Z_k^j, Z_k^i)\big), \\
\ell(Z_k^i, Z_k^j) = - \log \left( \frac{\exp\left( \text{sim}(Z_k^i, Z_k^j) / \tau \right)}{\sum_{n=1}^{N} \exp\left( \text{sim}(Z_n^i, Z_n^j) / \tau \right)} \right),
\end{split}
\end{equation}
where \( \text{sim}(\cdot) \) denotes the cosine similarity and \( \tau \) is the temperature parameter.

\subsubsection{Graph Clustering}

The clustering loss enhances representation clustering by minimizing the Kullback-Leibler (KL) divergence between soft clustering assignments and the target distribution. The similarity between the \(n\)-th row of fused representation \( Z_n \) and the cluster center \( \sigma_k \) is measured using Student's t-distribution:

\begin{equation}
q_{nk} = \frac{\left(1 + \| Z_n - \sigma_k \|^2 \right)^{-1}}{\sum_{j=1}^C \left( 1 + \| Z_n - \sigma_j \|^2 \right)^{-1}},
\end{equation}
where \( \sigma_k \) is  obtained by averaging the representations \(Z_n\) that share the same K-means pseudo-label, and the target distribution \( p \) is computed as:

\begin{equation}
p_{nk} = \frac{\left(q_{nk}\right)^2 / \sum_{n=1}^N q_{nk}}{\sum_{j=1}^C \left( \left(q_{nj}\right)^2 / \sum_{n=1}^N q_{nk} \right)}.
\end{equation}

The clustering loss minimizes the KL divergence between \( P \) and \( Q \):

\begin{equation}
\mathcal{L}_{CLU} = KL(P \,\|\, Q).
\end{equation}

\subsubsection{Overall Objective}

The total loss \( \mathcal{L} \), optimized via gradient descent, integrates reconstruction, contrastive, and clustering losses. The weights \( \lambda \) and \( \mu \) act as balancing factors, regulating the contributions of contrastive and cross-modality alignment losses:  

\begin{equation}  
\mathcal{L} = \mathcal{L}_{REC} + \lambda\mathcal{L}_{Cr} + \mu\mathcal{L}_{Cm} + \mathcal{L}_{CLU}.  
\end{equation}

\section{Experiment}

\subsection{Experimental Setups}
\textbf{Dataset} We conduct experiments on diverse types of datasets including two multimodal multi-relational graph datasets (Amazon\cite{jia2023multimodal} and IMDB\cite{jia2023multimodal}), three benchmark multi-relational graph datasets (ACM\cite{fu2020magnn}, Yelp\cite{lu2019relation} and DBLP\cite{yun2019graph}), and a large real-world multimodal graph Ele-fashion\cite{zhu2024multimodal}, which provide evidence for the broad applicability and scalability of our method. Details are shown in Table \ref{tab:dataset}.

\begin{table*}[h] 
\centering
\caption{Clustering performance comparison on Multimodal graph datasets. The highest values are marked in \textbf{bold}, and the second-highest values are \underline{underlined}. OOM indicates out-of-memory.}
\resizebox{\textwidth}{!}{ 
\renewcommand{\arraystretch}{0.9} 
\begin{tabular}{lccc|ccc|ccc}
\toprule
\multirow{2}{*}{Methods/Datasets} & \multicolumn{3}{c|}{IMDB} & \multicolumn{3}{c|}{Amazon} & \multicolumn{3}{c}{Ele-Fashion} \\
\cmidrule(lr){2-4} \cmidrule(lr){5-7} \cmidrule(lr){8-10}
& ACC\% & NMI\% & ARI\% & ACC\% & NMI\% & ARI\% & ACC\% & NMI\% & ARI\% \\
\midrule
KMeans-Text  & 50.61  & 12.28  & 14.58  & 61.19  & 15.29  & 21.36  & 35.97  & 34.84  & 25.48  \\
KMeans-Image  & 36.85  & 1.06   & -0.47  & 53.48  & 19.25  & 5.97   & 37.88  & 39.85  & 22.64  \\
\midrule
HDMI (WWW'21)        & 50.00  & 11.41  & 11.92  & 71.99  & 32.72  & 33.49  & 40.11  & 41.88  & 29.16  \\
InfoMGF-RA (NeurIPS'24)  & 51.45  & 12.83  & 15.22  & \underline{72.40}  & \underline{33.06}  & \underline{34.81}  & \underline{40.30}  & 47.36  & \underline{30.83}  \\
\midrule
GWMAC (CIKM ’22)      & 42.52  & 4.02   & 8.03   & 51.48  & 21.68  & 19.39  & 33.19  & 39.83  & 22.25  \\
GCFAgg (CVPR'23)      & \underline{53.20}  & \underline{13.43}  & \underline{16.79}  & 69.80  & 29.08  & 32.12  & 39.14  & \underline{51.69}  & 30.38  \\
\midrule
MHGAT (Neural Comput. Appl.'22)      & 40.53    & 1.40    & 0.14    & 64.53  & 11.45  & 16.71  & \multicolumn{3}{c}{OOM} \\
\midrule
DMGC        & \textbf{58.16} & \textbf{14.01} & \textbf{17.32} & \textbf{79.83} & \textbf{37.13} & \textbf{46.28} & \textbf{51.17} & \textbf{51.71} & \textbf{49.02} \\
\bottomrule
\end{tabular}
}
\label{tab:imdb_amazon_elefashion}
\end{table*}

\begin{table*}[h]
\centering
\caption{Clustering performance comparison on Multi-relational graph datasets.}
\resizebox{\textwidth}{!}{ 
\renewcommand{\arraystretch}{0.9} 
\begin{tabular}{lccc|ccc|ccc}
\toprule
\multirow{2}{*}{Methods/Datasets} & \multicolumn{3}{c|}{ACM} & \multicolumn{3}{c|}{DBLP} & \multicolumn{3}{c}{Yelp} \\
\cmidrule(lr){2-4} \cmidrule(lr){5-7} \cmidrule(lr){8-10}
 & ACC\% & NMI\% & ARI\% & ACC\% & NMI\% & ARI\% & ACC\% & NMI\% & ARI\% \\
\midrule
VGAE (NeurIPS'16)     & 73.58 & 45.07 & 43.47 & 84.48 & 61.79 & 65.56 & 65.07 & 39.19 & 42.57 \\
DGI  (ICLR'18)     & 81.14 & 57.79 & 51.74 & 86.88 & 65.59 & 70.35 & 65.29 & 39.42 & 42.62 \\
\midrule
O2MAC (WWW'20)    & 75.37 & 42.23 & 44.51 & 83.29 & 58.64 & 60.01 & 65.07 & 39.02 & 42.53 \\
MvAGC (TKDE'23)    & 59.49 & 18.19 & 18.79 & 78.39 & 50.39 & 51.21 & 63.14 & 24.39 & 29.25 \\
MCGC (NeurIPS'23)     & 71.29 & 53.07 & 43.96 & 87.96 & 65.56 & 71.51 & 65.61 & 38.35 & 35.17 \\
HDMI (WWW'21)     & 82.58 & 59.02 & 54.72 & 87.25 & 64.85 & 70.85 & 79.56 & 60.81 & 59.35 \\
MGDCR (TKDE'23)    & 68.38 & 54.47 & 43.72 & 81.91 & 62.47 & 62.22 & 72.71 & 44.23 & 46.47 \\
DMG (ICML'23)      & 87.96 & 63.41 & 67.26 & 88.45 & 69.03 & 73.07 & 88.26 & 65.66 & 66.33 \\
BTGF (AAAI'24)     & 88.53 & 64.83 & 67.76 & 88.05 & 66.28 & 72.47 & 91.39 & 69.97 & 73.53 \\
InfoMGF-RA (NeurIPS'24) & \underline{90.63} & \underline{68.11} & \underline{73.07} & \underline{88.72} & \underline{70.19} & \underline{73.49} & \underline{91.85} & \underline{72.67} & \underline{74.66} \\
DMGC      & \textbf{91.57} & \textbf{70.38} & \textbf{75.46} & \textbf{91.14} & \textbf{73.76} & \textbf{78.79} & \textbf{92.77} & \textbf{73.63} & \textbf{77.35} \\
\bottomrule
\end{tabular}
}
\label{tab:acm_dblp_yelp}
\end{table*}

\textbf{Baseline} For multimodal graph clustering task, we compare our approach with various baselines, including K-Means of raw textual and visual features, two multi-relational graph methods (HDMI\cite{jing2021hdmi} and InfoMGF-RA\cite{shen2024beyond}), two multimodal clustering methods (GWMAC \cite{10.1145/3511808.3557339} and GCFAgg\cite{yan2023gcfagg}), and a recent multimodal classification method (MHGAT\cite{jia2023multimodal}). For a fair comparison, we adapt MHGAT into an unsupervised setting by incorporating our proposed loss and for multi-relational graph methods, we evaluate each modality separately and report the highest result.

For multi-relational graph clustering task, we compare with two single-graph methods (i.e., VGAE\cite{kipf2016variationalgraphautoencoders} and DGI\cite{veličković2018deepgraphinfomax} ) and eight multi-relational graph methods (i.e., O2MAC\cite{fan2020one2multi}, MvAGC\cite{lin2023multi}, MCGC \cite{pan2021multi}, HDMI\cite{jing2021hdmi}, MGDCR\cite{mo2023multiplex}, DMG\cite{mo2023disentangled}, BTGF\cite{qian2024upper}, and InfoMGF-RA\cite{shen2024beyond}). We conduct single-graph methods separately for each graph and present the best results.

\textbf{Metrics} We adopt three popular clustering metrics\cite{10715731}, including Accuracy (ACC), Normalized Mutual Information (NMI), and Adjusted Rand Index (ARI). A higher value indicates a better performance.

\textbf{Parameter setting}  We conduct experiments with different learning rates ranging from \{1e-3, 2e-3, 3e-3, 5e-4\} and penalty weights for L2-norm regularization from \{1e-4, 5e-6, 0\}. Early stopping is used with patience of 50 epochs, stopping training if the total loss does not decrease for patience consecutive epochs. We set the number of homophilic graph neighbors \(k_l\) from \{10, 15, 20\}, while the number of heterophilic graph neighbors \(k_h\) from \{2, 3, 4, 5, 6\}. The number of layers is set from \{1, 2, 3, 5\}. The loss balancing weights \(\mu\) and \(\lambda\) are set from \{0, 1e-3, 1e-1, 1\}. All experiments are implemented in the PyTorch platform using an A40 48G GPU.

\subsection{Results Analysis}

Table \ref{tab:imdb_amazon_elefashion} reports the results in multimodal graph clustering. Multi-relational graph approaches (HDMI and InfoMGF-RA) fail to exploit multimodal information fully, leading to suboptimal results. Traditional multimodal methods (GCFAgg and GWMAC) lack graph structures, making them less effective for multimodal graph data, highlighting the importance of topology. Furthermore, transforming the supervised method (MHGAT) into an unsupervised one by adding the same loss performs poorly. In contrast, our approach achieves superior performance since it effectively integrates multimodal and topological information. 

Table \ref{tab:acm_dblp_yelp} presents the results in multi-relational graph clustering. Firstly, multi-relational graph clustering methods outperform single graph methods, demonstrating the advantages of leveraging information from multiple sources. Secondly, graph structure learning methods generally display better outcomes than the others. This is because the original graph structure contains task-independent noise that affects the results of the unsupervised task, which is consistent with the findings from our empirical analysis. Finally, DMGC achieves notably superior results, owing to the consideration of graph structure learning and combining dual-channel information.

\begin{figure}[t]
    \centering
    \vspace{-1.0em}
    \begin{minipage}{0.36\linewidth}
        \centering
        \subfloat[Raw: 0.0679]{\includegraphics[width=\linewidth]{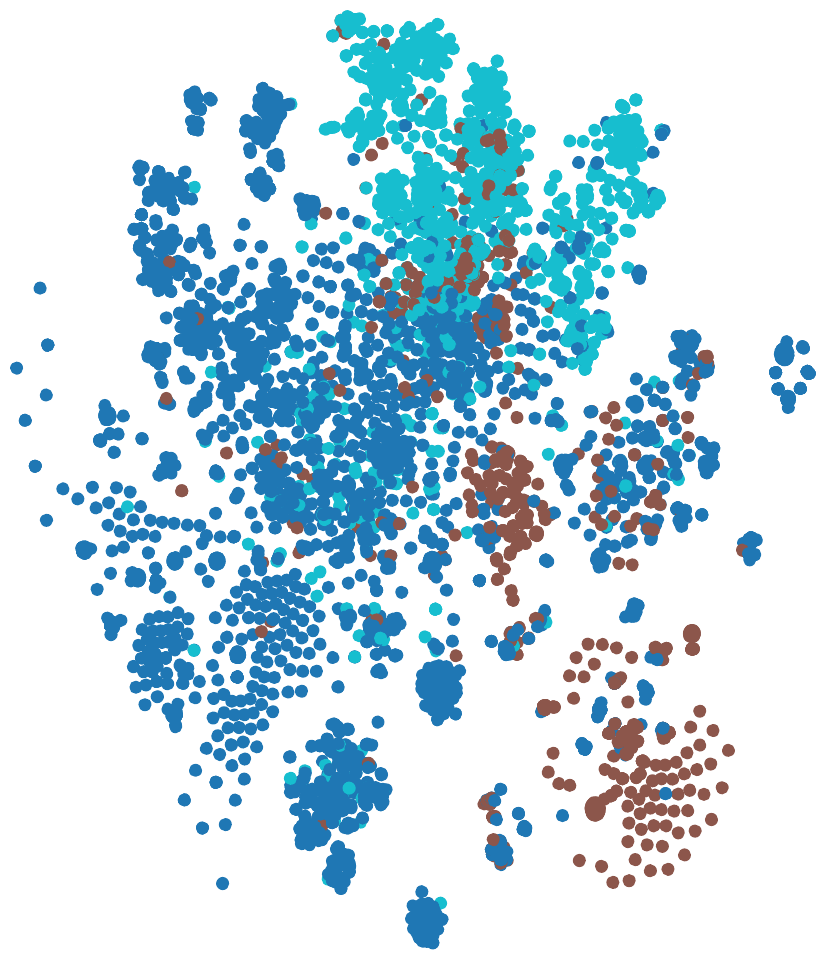}\label{Raw}} 
    \end{minipage}
    \hspace{2em} 
    \begin{minipage}{0.36\linewidth}
        \centering
        \subfloat[InfoMGF-RA: 0.1088]{\includegraphics[width=\linewidth]{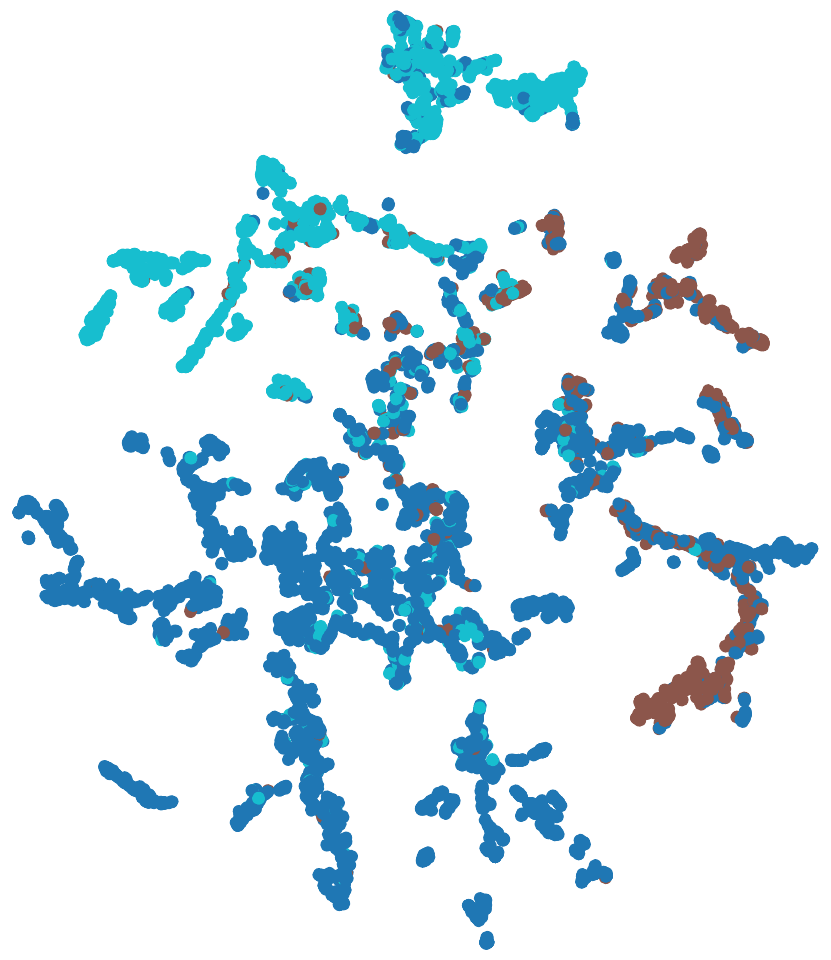}\label{InfoMGF-RA}}
    \end{minipage}
    
    \vspace{1em} 

    \begin{minipage}{0.36\linewidth}
        \centering
        \subfloat[GCFAgg: 0.1425]{\includegraphics[width=\linewidth]{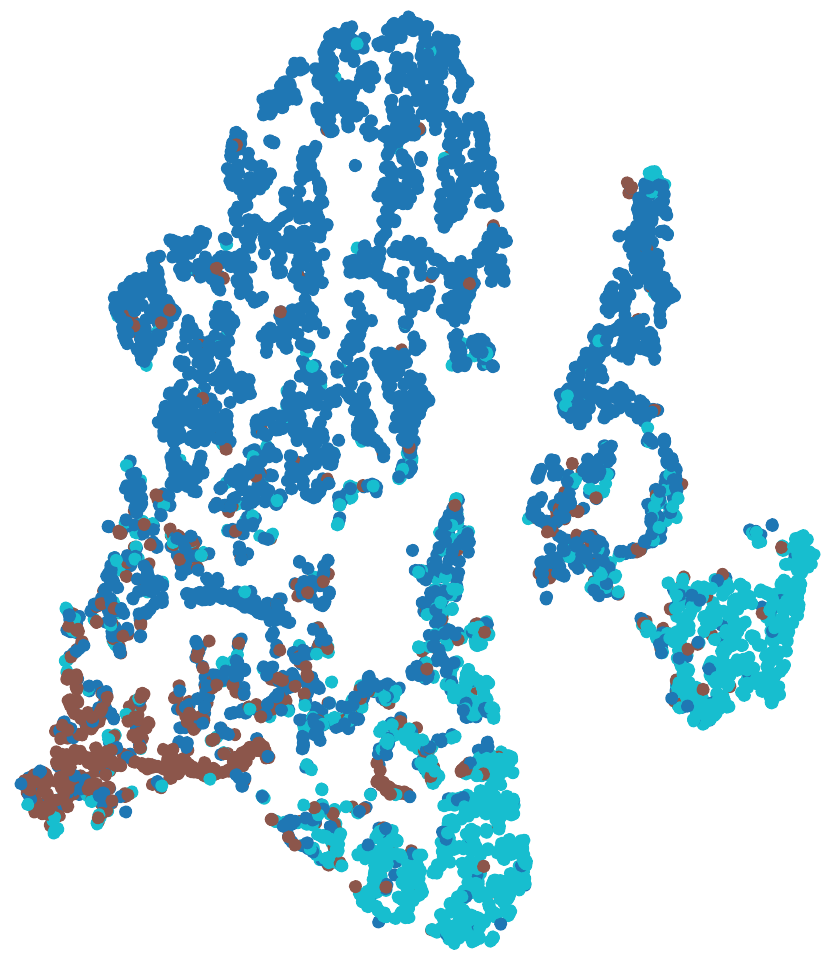}\label{GCFAgg}}
    \end{minipage}
    \hspace{2em} 
    \begin{minipage}{0.36\linewidth}
        \centering
        \subfloat[DMGC: 0.2200]{\includegraphics[width=\linewidth]{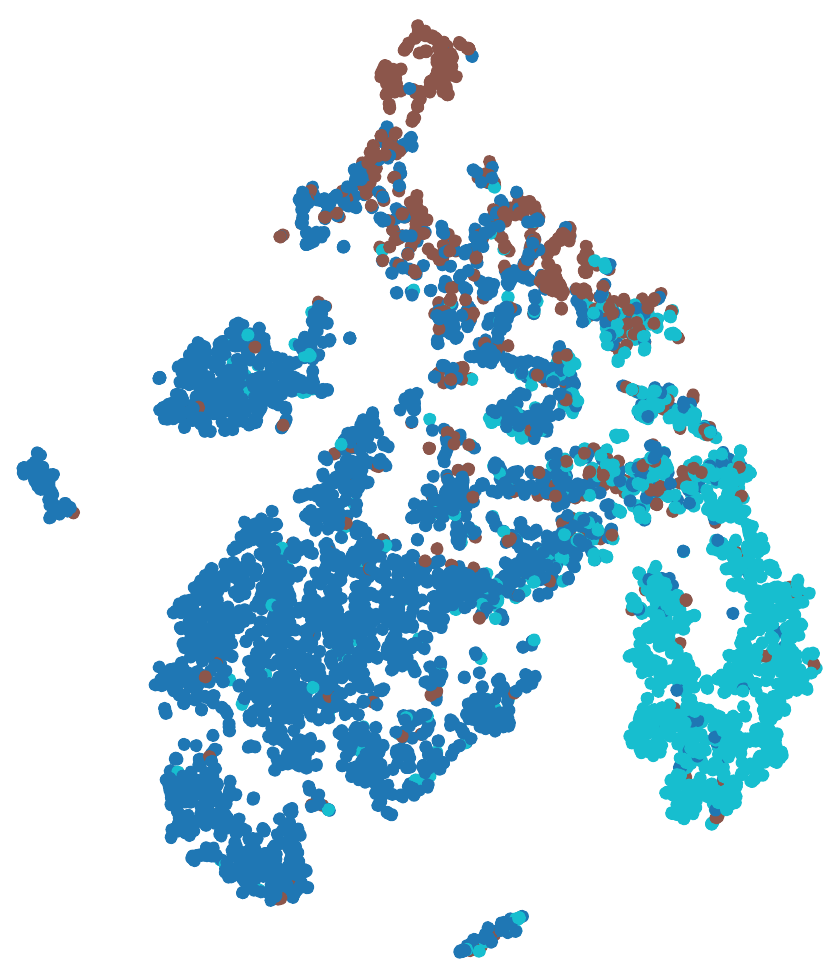}\label{Ours}}
    \end{minipage}
    \vspace{-0.5em}
    \caption{Visualization of the learned node representation on Amazon. The corresponding Silhouette scores are also given.}
    \vspace{-1.0em}
    \label{Visual}
\end{figure}

To provide an intuitive evaluation, we visualize learned node representations on the Amazon dataset through t-SNE in Figure \ref{Visual}. Different colors indicate different classes. We could observe that the raw features exhibit significant inter-class mixing with poorly separated category distributions. While InfoMGF-RA and GCFAgg demonstrate partial separation, considerable inter-class overlap persists. In contrast, our method achieves superior class discrimination through: (1) noticeably reduced overlap regions and (2) significantly tighter concentration of correct-category nodes around cluster centers, though perfect separation remains challenging in this feature space. Furthermore, we calculate the Silhouette score for the generated clusters, with higher values reflecting improved clustering quality. Our method outperforms all other approaches, highlighting its effectiveness and the strong category discrimination achieved by the learned representations.

\subsection{Ablation Study}


\textbf{Effectiveness of loss components} To verify the effectiveness of the reconstruction loss and contrastive learning loss, we design three variations (w/o $\mathcal{L}_r$, w/o $\mathcal{L}_{Cr}$ and w/o $\mathcal{L}_{Cm}$) in Table \ref{ablation}. Notably, modality-specific reconstruction loss preserves unique modal features, frequency domain alignment harmonizes high- and low-frequency components of fused representations, and modality alignment ensures inter-modality coherence. Each component is indispensable, as the absence of any would significantly compromise the robustness and effectiveness of the learned representations.

\begin{table*}[t]
    \centering
    \begin{minipage}{0.75\linewidth} 
        \centering
        \captionsetup{type=table}
        \caption{Performance of Ours and its variants.}
        \resizebox{\linewidth}{!}{
            \renewcommand{\arraystretch}{0.9} 
            \begin{tabular}{c|cc|cc|cc|cc|cc}
                \toprule
                \multirow{2}{*}{Variants} & \multicolumn{2}{c|}{IMDB} & \multicolumn{2}{c|}{Amazon} & \multicolumn{2}{c|}{ACM} & \multicolumn{2}{c|}{DBLP} & \multicolumn{2}{c}{Yelp} \\    
                & NMI    & ACC   & NMI    & ACC   & NMI    & ACC   & NMI    & ACC   & NMI    & ACC   \\
                \midrule
                w/o $\mathcal{L}_r$       & 12.01 & 53.38 & 23.42 & 72.29 & 67.19 & 89.51 & 68.83 & 85.37 & 71.77 & 92.10 \\
                w/o $\mathcal{L}_{Cr}$       & 9.09 & 52.86 & 29.32 & 76.25 & 56.49 & 71.44 & 68.96 & 86.57 & 64.59 & 82.22 \\
                w/o $\mathcal{L}_{Cm}$   & 12.66 & 54.88 & 25.90 & 63.48 & --- & --- & --- & --- & --- & --- \\
                \midrule
                w/o Homo.                 & 10.56  & 52.89 & 21.01 & 70.91 & 56.29  & 70.42 & 67.84 & 86.49 & 67.31 & 89.76 \\
                w/o Heter.                & 11.11 & 46.72 & 28.26 & 70.04  & 65.72 & 87.40 & 70.80 & 89.45 & 69.47 & 89.22 \\
                \midrule
                Ours                      & 14.01 & 58.16 & 37.13 & 79.83 & 70.38 & 91.57 & 73.76 & 91.14 & 73.63  & 92.77 \\
                \bottomrule
            \end{tabular}
        }
        \label{ablation}
    \end{minipage}
\end{table*}

\textbf{Effectiveness of homophilic and heterophilic graphs} Our method successfully decouples hybrid neighborhood patterns into homophilic and heterophilic graphs for multimodal multi-relational learning. Through ablation studies with two variants - one excluding homophilic structures (w/o Homo) and another excluding heterophilic structures (w/o Heter) - we demonstrate the critical importance of this disentanglement. The results reveal that while homophilic structures play a dominant role (evidenced by greater performance degradation when removed), heterophilic graphs provide unique value in capturing complementary relational patterns and enriching modality-specific representations. This empirical validation confirms that our approach to explicitly modeling and combining both graph types is essential for robust learning, as each captures distinct but complementary aspects of the underlying hybrid neighborhood patterns.

\begin{figure}[t]
    \centering
    \includegraphics[width=0.9\linewidth]{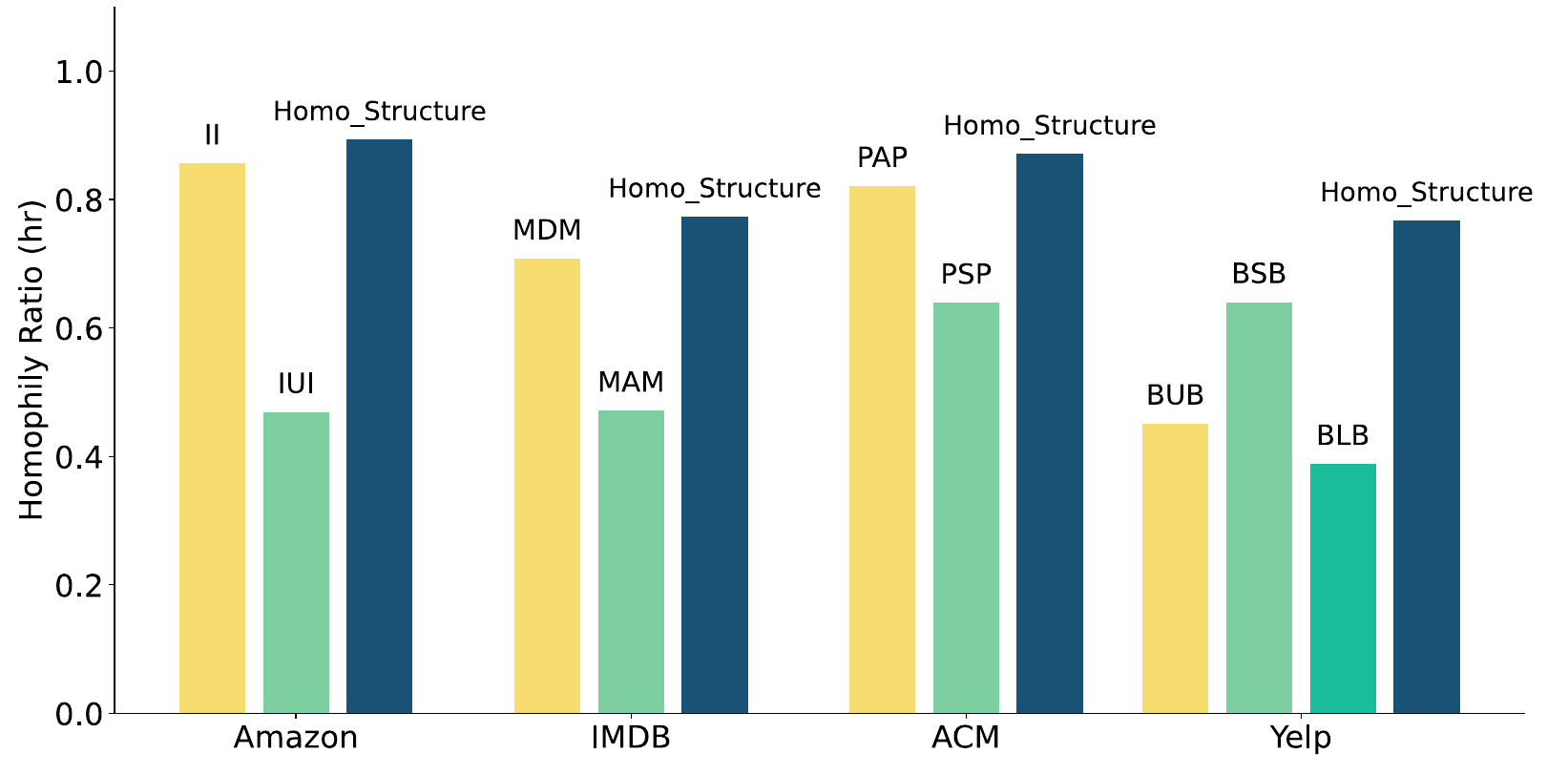}
    \caption{Homophily ratio comparison of different datasets.}
    \label{fig:hrc}
    \vspace{-2.3em}
\end{figure}

\textbf{Effectiveness of homophily ratio improvement} To demonstrate that the multimodal multi-relational fusion unified graph generated by our method achieves a higher homophily ratio, we compare the homophily ratios of bipartite graphs constructed from multiple datasets with those of the original adjacency matrices, as illustrated in Figure \ref{fig:hrc}. The results exhibit a consistent improvement across all datasets, validating that our method effectively enhances structural coherence and preserves meaningful relationships, leading to the construction of high-quality matching graphs.

\subsection{Sensitivity Analysis} 

\begin{figure}[h]
    \vspace{-0.8em}
    \centering
    \includegraphics[width=0.7\linewidth]{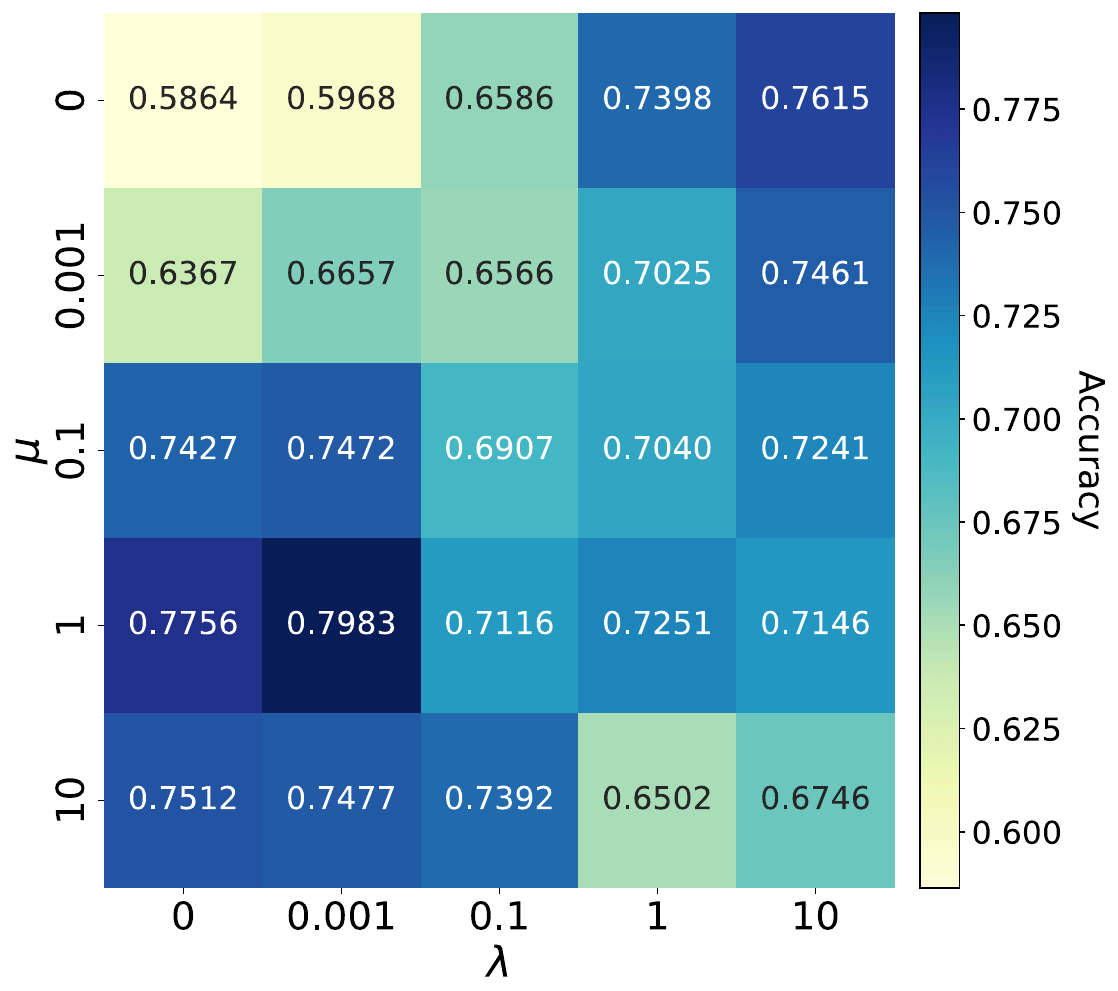}
    \caption{Accuracy heatmap of Amazon between loss balancing hyperparameters}
    \label{fig:hm}
    \vspace{-2.0em}
\end{figure}

\textbf{Loss balancing hyperparameters analysis.} Figure \ref{fig:hm} shows the heatmap on Amazon, illustrating the sensitivity of the model's performance concerning two hyperparameters: $\lambda$, the coefficient for intra-modality low-pass and high-pass alignment loss, and $\mu$, the coefficient for cross-modality alignment loss. We note that increasing $\lambda$ generally improves performance when $\mu$ is small (e.g., $\mu = 0$), reaching its peak at $\lambda = 10$. However, for larger $\mu$ (e.g., $\mu = 1$ or $10$), an excessive $\lambda$ leads to performance degradation, suggesting that overly strong intra-modality alignment might hinder generalization. On the other hand, increasing $\mu$ from $0$ to $1$ consistently enhances performance, indicating the effectiveness of cross-modality alignment. The highest performance is achieved at $\lambda = 0.001$ and $\mu = 1$, where the score reaches $0.7983$. However, an excessively large $\mu$ (e.g., $\mu = 10$) tends to decrease performance across different $\lambda$ values, implying that excessive cross-modality alignment may suppress intra-modality variations. Overall, the results suggest that a balanced choice of $\lambda$ and $\mu$ is crucial for optimal performance, with moderate $\lambda$ (e.g., $0.001$) and properly tuned $\mu$ (e.g., $1$) yielding the best results.


\begin{figure}[t]
    \centering
    \begin{minipage}{0.78\linewidth}
        \centering
        \includegraphics[width=\linewidth]{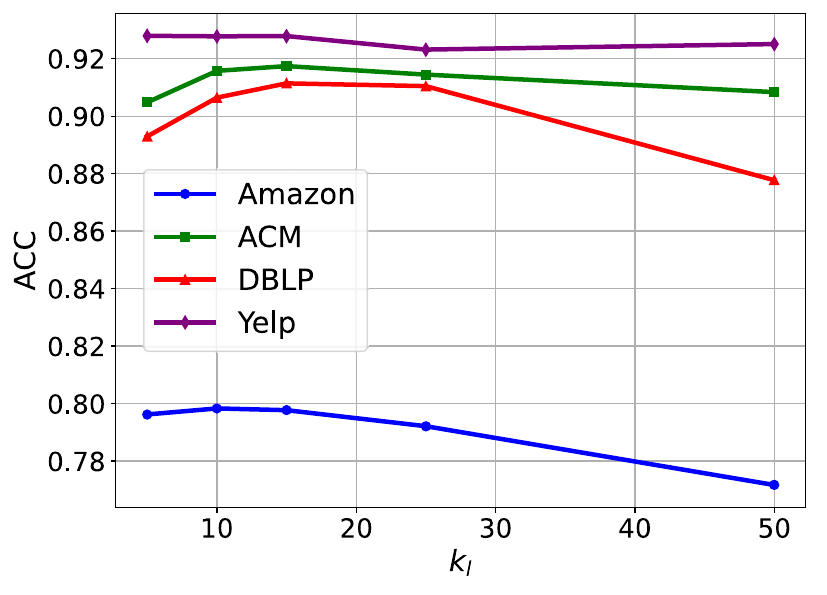}
        \vspace{-2.0em}
        \caption{Sensitivity of \(k_l\) in Homophily Graph Construction}
        \label{fig:hrl}
    \end{minipage}
    \hfill
    \begin{minipage}{0.78\linewidth}
        \centering
        \includegraphics[width=\linewidth]{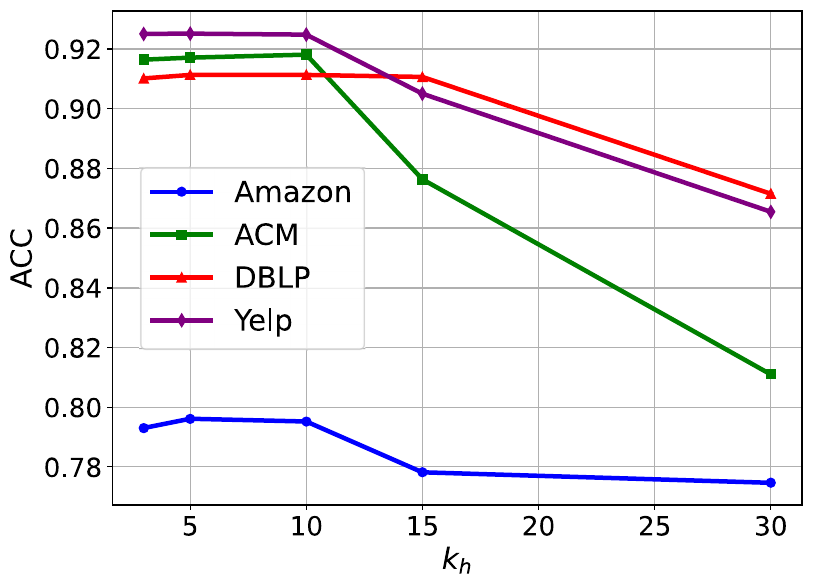}
        \vspace{-2.0em}
        \caption{Sensitivity of \(k_h\) in Heterophily Graph Construction}
        \label{fig:hrh}
    \end{minipage}
    \vspace{-2.3em}
\end{figure}

\textbf{Hyperparameter Analysis in \(K\)-Nearest Neighbor Graph Construction.} Figures \ref{fig:hrl} and \ref{fig:hrh} illustrate the effect of hyperparameters \( k_l \) and \( k_h \) for ACC in $k$-nearest neighbor ($k$NN) graph construction, where \( k_l \) and \( k_h \) control the number of neighbors in the homophilic and heterophilic graphs, respectively. As \( k \) increases, the ACC tends to decline, especially in the heterophilic graph, since an excessively large \( k_l \) or \( k_h \) hampers the disentanglement of high-quality homophilic and heterophilic signals from the hybrid neighborhood pattern. This degrades graph quality by mixing information from different frequency domains, ultimately affecting model performance. 

In contrast, when \( k \) is small, the ACC remains almost stable, demonstrating the robustness of our graph construction method. These results highlight the critical role of properly selecting \( k_l \) and \( k_h \) to ensure high-quality graph representations and optimal clustering performance.

\section{Conclusion}
In this paper, we propose DMGC, the first comprehensive framework for unsupervised multimodal graph clustering, tackling hybrid neighborhood patterns, multi-relational complexity, and multimodal feature fusion. Experiments on six real-world datasets confirm its effectiveness, achieving state-of-the-art results and scaling to graphs with up to 97K nodes. This work lays a foundation for future research in unsupervised multimodal graph learning and aims to inspire further progress in the field.

\section*{Acknowledgements}

This work was supported by the National Natural Science Foundation of China (No. U24A20323).

\bibliographystyle{plain}
\bibliography{reference.bib}

\appendix


\section{Additional Visualization Details}

\subsection{T-SNE visualization comparison between homophilic and heterophiic graph embeddings}

We conducted an additional experiment on the unimodal dataset ACM to provide a visualization comparison between homophilic and heterophilic graph embeddings. From Figure \ref{fig:combined}, we have two key observations: 

\begin{figure}[h]
    \centering
    \begin{subfigure}[b]{0.48\linewidth}
        \includegraphics[width=\linewidth]{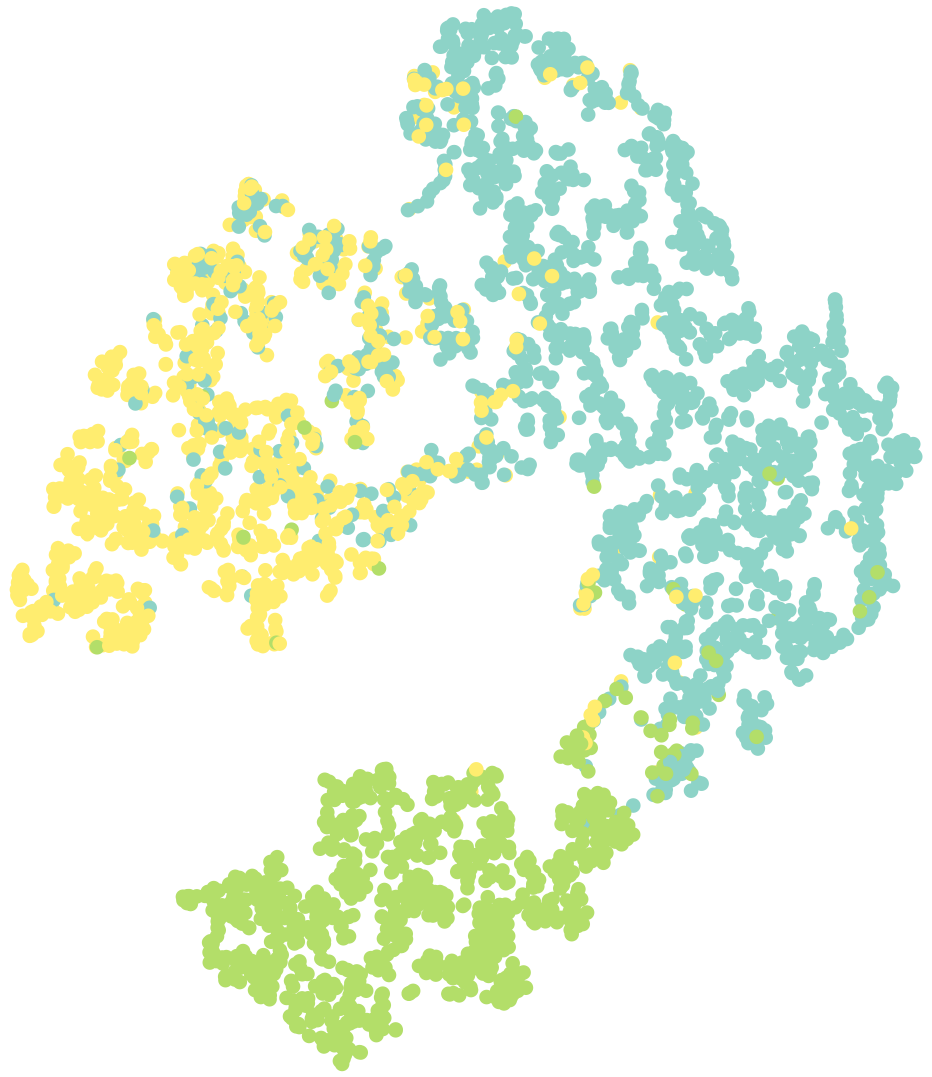}
        \caption{Homophilic graph embeddings $Z_l$ of dataset ACM}
        \label{z_l_tsne}
    \end{subfigure}
    \hfill
    \begin{subfigure}[b]{0.48\linewidth}
        \includegraphics[width=\linewidth]{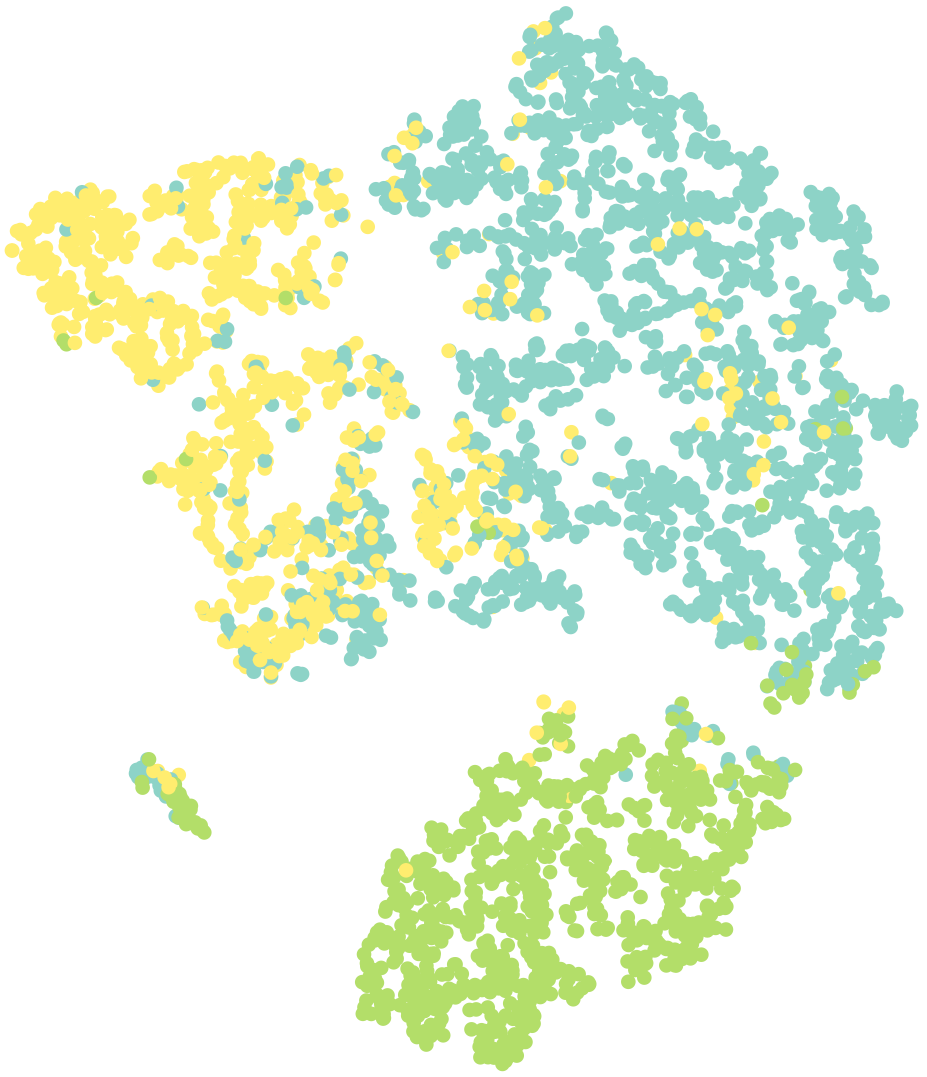}
        \caption{Heterophilic graph embeddings $Z_h$ of dataset ACM}
        \label{fig:sub2}
    \end{subfigure}
    \caption{T-SNE visualization comparison between homophilic and heterophilic graph embeddings of dataset ACM}
    \label{fig:combined}
\end{figure}

Homophilic embeddings form dense clusters with blurred inter-class boundaries, which is consistent with their low-pass filtering behavior that emphasizes shared features. Heterophilic embeddings exhibit inter-class dispersion, where nodes from different classes overlap in the embedding space, reflecting high-frequency signal preservation. They also demonstrate intra-class separation, with structural outliers within the same class being distinguishable, which validates their high-pass filtering effect. This visualization empirically justifies our dual-filter design, where homophily captures cross-modality consistency, while heterophily emphasizes modal-specific distinctiveness.

\subsection{Visualization of the homophily-enhanced graph and heterophily-aware graph}
We provide an intuitive visualization of the homophily-enhanced and heterophily-aware graphs learned by DMGC on the Yelp dataset. We construct class-wise connection matrices based on the adjacency structure and node labels, where each entry $(i,j)$ indicates the number of edges between class $i$ and class $j$.

\begin{figure}[t]
    \centering
    \begin{subfigure}[t]{0.39\textwidth}
        \includegraphics[width=\linewidth]{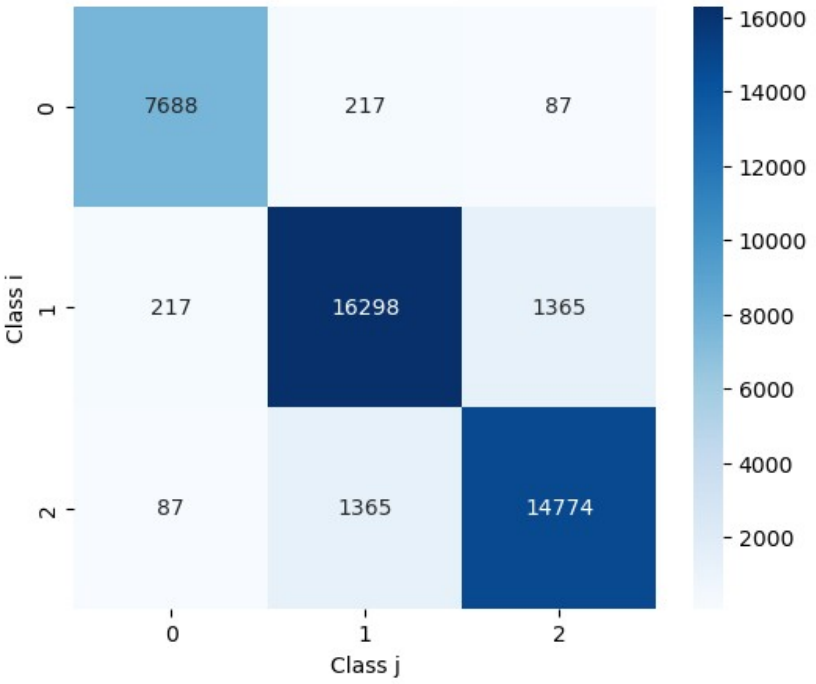}
        \caption{Class-wise connection matrix for homophily-enhanced graph}
        \label{fig:homo}
    \end{subfigure}
    \hfill
    \begin{subfigure}[t]{0.39\textwidth}
        \includegraphics[width=\linewidth]{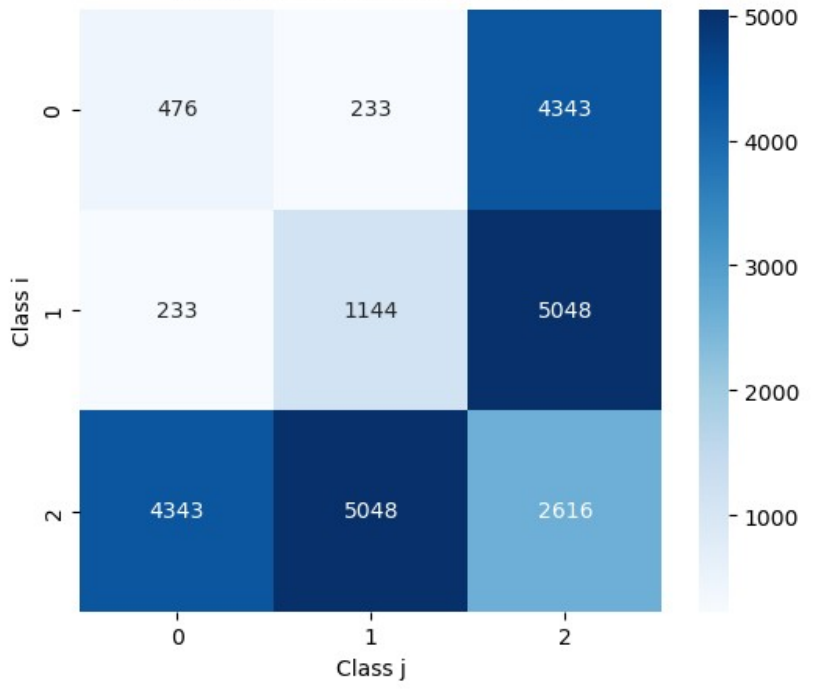}
        \caption{Class-wise connection matrix for heterophily-aware graph}
        \label{fig:hetero}
    \end{subfigure}
    \caption{Class-wise connection matrices visualized from the graphs learned by DMGC.}
    \label{fig:homo_hetero}
\end{figure}

Figure~\ref{fig:homo_hetero} shows that the homophily-enhanced graph exhibits strong diagonal patterns, reinforcing intra-class connections, while the heterophily-aware graph highlights off-diagonal patterns, enhancing inter-class connections.

\renewcommand{\floatpagefraction}{.8} 
\begin{table*}[ht!]
\centering
\caption{Statistics of the datasets.}
\resizebox{\linewidth}{!}
{
\renewcommand{\arraystretch}{1.25}
\begin{tabular}{|c|c|c|c|c|}
\hline
\textbf{Dataset} & \textbf{Relation}                                                                                                     & \textbf{Modality}                                    & \textbf{Category} & \textbf{Node} \\ \hline
IMDB             & \begin{tabular}[c]{@{}c@{}}Movie-Actor-Movie\\ Movie-Director-Movie\end{tabular}                                      & \begin{tabular}[c]{@{}c@{}}Text\\ Image\end{tabular} & 3                 & 4278          \\ \hline
Amazon           & \begin{tabular}[c]{@{}c@{}}Item-User-Item\\ Item-Item\end{tabular}                                                    & \begin{tabular}[c]{@{}c@{}}Text\\ Image\end{tabular} & 3                 & 6158          \\ \hline
ACM              & \begin{tabular}[c]{@{}c@{}}Paper-Author-Paper\\ Paper-Subject-Paper\end{tabular}                                      & Text                                                 & 3                 & 4019          \\ \hline
DBLP             & \begin{tabular}[c]{@{}c@{}}Author-Paper-Author\\ Author-Paper-Conference-Paper-Author\end{tabular}                    & Text                                                 & 4                 & 2957          \\ \hline
Yelp             & \begin{tabular}[c]{@{}c@{}}Business-Service-Business\\ Business-User-Business\\ Business-Rating-Business\end{tabular} & Text                                                 & 3                 & 2614          \\ \hline
Ele-Fashion      & Item-Item                                                                                                             & \begin{tabular}[c]{@{}c@{}}Text\\ Image\end{tabular} & 11                & 97,766        \\ \hline
\end{tabular}
}
\label{tab:dataset}
\end{table*}

\section{Additional Experiments}

We have evaluated our model on two real-world heterophilic graph datasets (Texas and Chameleon), which are both sparse (The adjacency matrix exhibits a density of less than 3\%) and challenging (The adjacency matrices of the Texas and Chameleon datasets exhibit low homophily ratios of 0.09 and 0.23, respectively). Our method achieves strong performance, specifically, on Texas, it attains 37.1\% NMI\% and 62.9\% ACC\%, while on Chameleon, it reaches 22.5\% NMI\% and 39.9\% ACC\%.
Additionally, robustness is verified on Amazon dataset with 20\% Gaussian noise, where accuracy drops by $\leq$ 4.7\%. This demonstrates that our method not only enables precise semantic matching through contrastive feature purification but also achieves superior generalization by going beyond trivial feature correlations.

\section{Statistics of the datasets}

Table \ref{tab:dataset} summarizes the basic information of the six datasets used in the experiment, including relation, modality, category, and node.

\end{document}